\documentclass[letterpaper]{article} 
\usepackage{aaai24}  
\usepackage{times}  
\usepackage{helvet}  
\usepackage{courier}  
\usepackage[hyphens]{url}  
\usepackage{graphicx} 
\usepackage{natbib}  
\usepackage{caption} 
\usepackage{algorithm}
\usepackage{algorithmic}
\usepackage{amsmath}
\usepackage{xcolor}   
\usepackage{subcaption}
\usepackage{float}

\usepackage{newfloat}
\usepackage{listings}
\DeclareCaptionStyle{ruled}{labelfont=normalfont,labelsep=colon,strut=off} 
\floatstyle{ruled}
\newfloat{listing}{tb}{lst}{}
\floatname{listing}{Listing}
\title{Word2World: Generating Stories and Worlds through Large Language Models}
\author {
    Muhammad Umair Nasir,\textsuperscript{\rm 1}
    Steven James,\textsuperscript{\rm 1}
    Julian Togelius \textsuperscript{\rm 2}
}
\affiliations {
    \textsuperscript{\rm 1} University of the Witwatersrand, Johannesburg, South Africa\\
    \textsuperscript{\rm 2} New York University, New York, USA\\
    umairnasir1@students.wits.ac.za, steven.james@wits.ac.za, julian@togelius.com
}
\usepackage{bibentry}

\begin{document}

\maketitle

\begin{abstract}
Large Language Models (LLMs) have proven their worth across a diverse spectrum of disciplines. LLMs have shown great potential in Procedural Content Generation (PCG) as well, but directly generating a level through a pre-trained LLM is still challenging. This work introduces \texttt{Word2World}, a system that enables LLMs to procedurally design playable games through stories, without any task-specific fine-tuning. \texttt{Word2World} leverages the abilities of LLMs to create diverse content and extract information. Combining these abilities, LLMs can create a story for the game, design narrative, and place tiles in appropriate places to create coherent worlds and playable games. We test \texttt{Word2World} with different LLMs and perform a thorough ablation study to validate each step. We open-source the code at \url{https://github.com/umair-nasir14/Word2World}.
\end{abstract}

\newcommand{\TODO}[1]{{\color{red} TODO: {#1}}}

\section{Introduction}

Generating content for games and virtual worlds is an important challenge as virtual environment are not only becoming more widespread and integral to the fabric of society, but also larger and more costly to develop. While algorithmic solutions for content generation have existed for a long time, they have typically been very narrow in terms of the type of content they can generate, or the environments they support. For example, there are methods that generate full levels for single games, or specific artifacts like trees and bushes across game environments. Generating content across multiple facets or games, such as maps, characters, and narrative, as well as generating content across games, are largely open research questions. Progress on these questions has the potential not only to advance tooling for designers of games and virtual worlds, but also to enable player-adaptive games and tools for users to generate their own content. Here, we specifically address the challenge of jointly generating narratives, characters, and playable world maps.

We achieve above-mentioned challenge through Large Language Models (LLMs). Artificial intelligence has been profoundly transformed by the invention of LLMs, which have demonstrated remarkable versatility across many domains. These models, exemplified by their ability to process and generate text in a manner that is contextually and semantically rich, have not only redefined the boundaries of natural language processing~\cite{liu2023pre} but have also shown unprecedented versatility in tasks far beyond text generation, such as generating neural networks~\cite{nasir2023llmatic} and folding proteins~\cite{lin2022language}. Among the frontier challenges in AI, the generation of 2D game levels represents a unique intersection of creativity and algorithmic complexity, offering a fertile ground for exploring the capabilities of LLMs beyond their conventional applications. LLMs have also achieved success in generating game levels~\cite{todd2023level,sudhakaran2024mariogpt}.

LLMs' potential to revolutionise the level generation field lies in their ability to understand and generate complex structures and patterns, a prerequisite for creating engaging and challenging game environments. Traditionally, game-level generation has relied on a combination of human creativity and special-purpose procedural content generation algorithms~\cite{shaker2016procedural}. However, the integration of LLMs, particularly in a zero-shot context, presents an intriguing possibility: the ability to generate novel game levels without the need for extensive task-specific training or fine-tuning. This zero-shot capability is predicated on the models' ability to generalize from a vast corpus of learned data to new, unseen tasks~~\cite{brown2020language}.

\begin{figure*}
    \centering
    \includegraphics[width=1.0\linewidth]{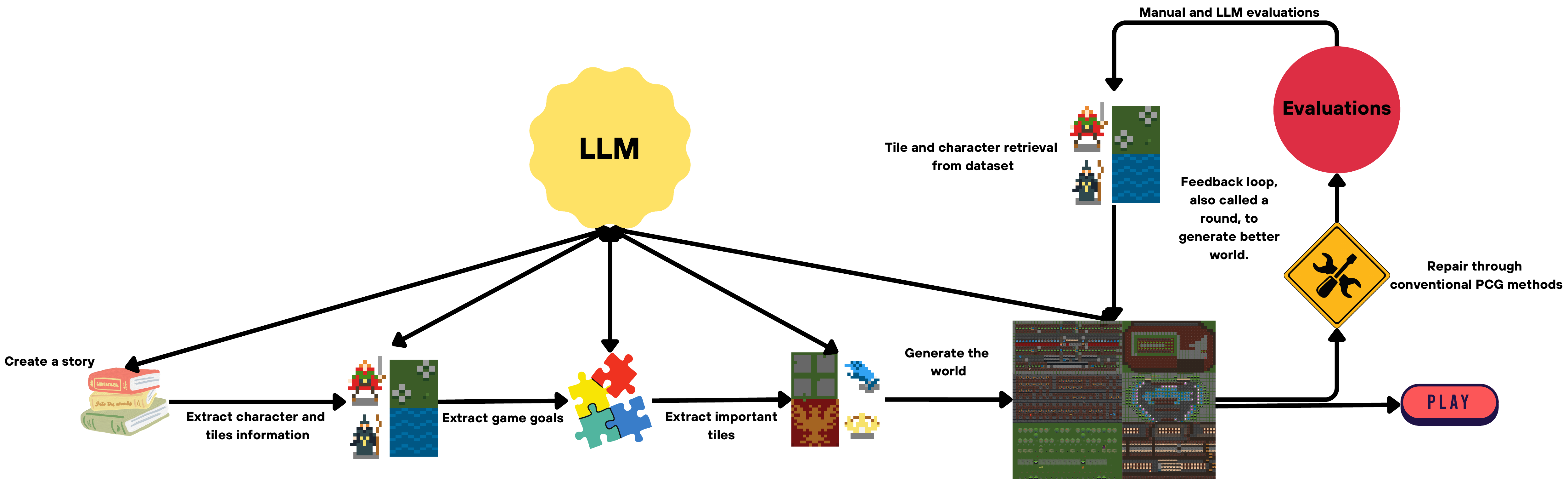}
  \caption{A flowchart of the whole pipeline for \texttt{Word2World}. An LLM is required for each step. Once all the required information is extracted, we move to a feedback loop called rounds. This loop makes sure the end result is a playable game that is coherent with the story.}
    \label{fig:flowchart}
\end{figure*}

LLMs showcase impressive zero-shot learning capabilities, yet they encounter significant challenges when tasked with generating a 2D game level directly from a text prompt~\cite{nasir2023practical}. This complexity stems from the nuanced requirements of game-level design, where elements like the narrative of the game, environment design, character placement, pathway clarity, and objective definition are crucial for creating a coherent and engaging player experience. For instance, ensuring that protagonists and antagonists are strategically positioned, designing navigable paths that guide players while maintaining an element of challenge, and articulating clear objectives are just some of the intricate details that must be meticulously considered. These requirements are essential not only for gameplay but also for sustaining narrative depth and visual coherence, which are pivotal in immersing players in the game world. Thus, automating the generation of a playable 2D game level through text, without any specialised fine-tuning, poses a significant research problem. 

Overcoming this challenge could revolutionize game design, enabling the rapid production of diverse and innovative game levels, and thereby expanding the creativity of the gaming industry. Moreover, it will allow the research community to progress immensely in computational game creativity \cite{liapis2014computational}, as every step in game design requires creativity, such as creating a story and converting it into a narrative by creating how each character looks, how the environment of the world looks like, and what are the objectives of the game. Furthermore, computational game creativity then includes the knowledge of what tiles to use and how should we design the aesthetic world and which should be coherent with the story. A system that can generate all of this content, while being coherent to the story and aesthetically pleasing to the players is eventually a very hard task.

Thus, this research introduces \texttt{Word2World}, a novel methodology that leverages the capabilities of LLMs to convert story into a narrative and into playable 2D game levels, marking a significant advancement in the domain of zero-shot LLM capabilities. This study aims to develop a system where LLMs can interpret storytelling elements, and transform them into coherent and engaging game environments by extracting useful information like what are the character and tiles descriptions, and what should be the goals of the player to complete the game. This approach involves the models' understanding and integrating narrative structures, character roles, and plot-driven events into game-level design. The research evaluates how LLMs can blend story elements with game design principles, and manage complexity within the constraints of 2D game environments. This method not only broadens the applications of LLMs but also provides a new perspective on the relationship between storytelling and game design, potentially establishing new benchmarks for narrative-driven game development and offering pathways for creating more immersive and personalized gaming experiences.

\section{Word2World}

In this section, we will discuss what \texttt{Word2World} is and how it works. \texttt{Word2World} is a system that can generate playable games (or environments) from stories. Generation of playable games through text prompts is a hard task, thus direct generation through prompts is very unlikely to give desired results. \texttt{Word2World} creates levels in steps. Each step is important and discussed in detail in this section. The steps are illustrated in Figure \ref{fig:flowchart} and Algorithm \ref{alg:sw}.

\subsection{Procedural Content Generation}

At the centre of the architecture, we have an LLM. An instance of the LLM generates a story (Line 1 of Algorithm \ref{alg:sw}), that is the basis of the whole game. From this story, the LLM then extracts character information, tileset information, goals, important tiles, walkable tiles, and interactive tiles (Line 2-7 of Algorithm \ref{alg:sw}). Then we provide this information to the LLM and ask it to generate a 2D world as alphanumeric characters, in 2 steps (Line 14-15 of Algorithm \ref{alg:sw}). The first step lays down the world's environment, while the second lays down characters and important interactive tiles. In the upcoming sections, we will look into detailed ablation studies and explain why 2 steps are necessary. We then use algorithmic checks to see if the placement of the number of character tiles is correct and if each row of the tile map is the same size (Line 16 of Algorithm \ref{alg:sw}). If the number of similar characters is more than one then we delete all except one that is first from the top-left side. If the rows are not equal then we pad the rows to the maximum length found among all rows. The rows are padded with the last character found in the row, if it is a game character then we pad with the previous character. 

Once we have our world laid down, we move to evaluating the world (Line 17 of Algorithm \ref{alg:sw}). The evaluations are done by the LLM as an evaluator and the conventional PCG methods, such as finding paths through AStar agent. The evaluations are stored for the next round. All rounds, except the first one, get previous world and evaluation scores as feedback to improve the world(Line 8-17 of Algorithm \ref{alg:sw}). For tile, we use two datasets, first for the environment and interactive objects and the second for characters. We retrieve tiles and create the world (Line 19-20 of Algorithm \ref{alg:sw}).

\begin{algorithm}
\caption{\texttt{Word2World}}
\label{alg:sw}
\begin{algorithmic}[1]
\REQUIRE $LLM\;for\;each\;step$
\STATE $S = create\_story()$
\STATE $C_I = extract\_character\_information(S)$
\STATE $T_I = extract\_tileset\_information(S, C_I)$
\STATE $G = extract\_goals(S, C_I, T_I)$
\STATE $I_T = extract\_important\_tiles(S, C_I, T_I, G)$
\STATE $W_T = extract\_walkable\_tiles(S, C_I, T_I, G)$
\STATE $O_T=extract\_object\_tiles(S, C_I, T_I, G)$

\FOR{$i \leftarrow total\_rounds$}
    \IF{$i > 0$}
        \STATE $Evals\;and\;W\;as\;feedback\;for\;generation$
    \ELSE
        \STATE $No\;feedback\;for\;world\;generation$
        
    \ENDIF
    \STATE $W_E=world\_environment(S, C_I, T_I, I_T,W_T,O_T)$
    \STATE $W=generate\_world(S, C_I, T_I, tiles, W_E)$
    \STATE $W=algorithmic\_fixes(W)$
    \STATE $Evals=evaluations(W)$   
\ENDFOR
\STATE $tiles=similarity\_score(T_I, tile\_dataset)$
\STATE $Generated\;world=Paste\_tiles(tiles,W)$
\RETURN $Generated\;world$
\end{algorithmic}
\end{algorithm}

\subsection{Tile Selection}

\begin{figure}[ht!]
    \centering
    \includegraphics[width=0.5\linewidth]{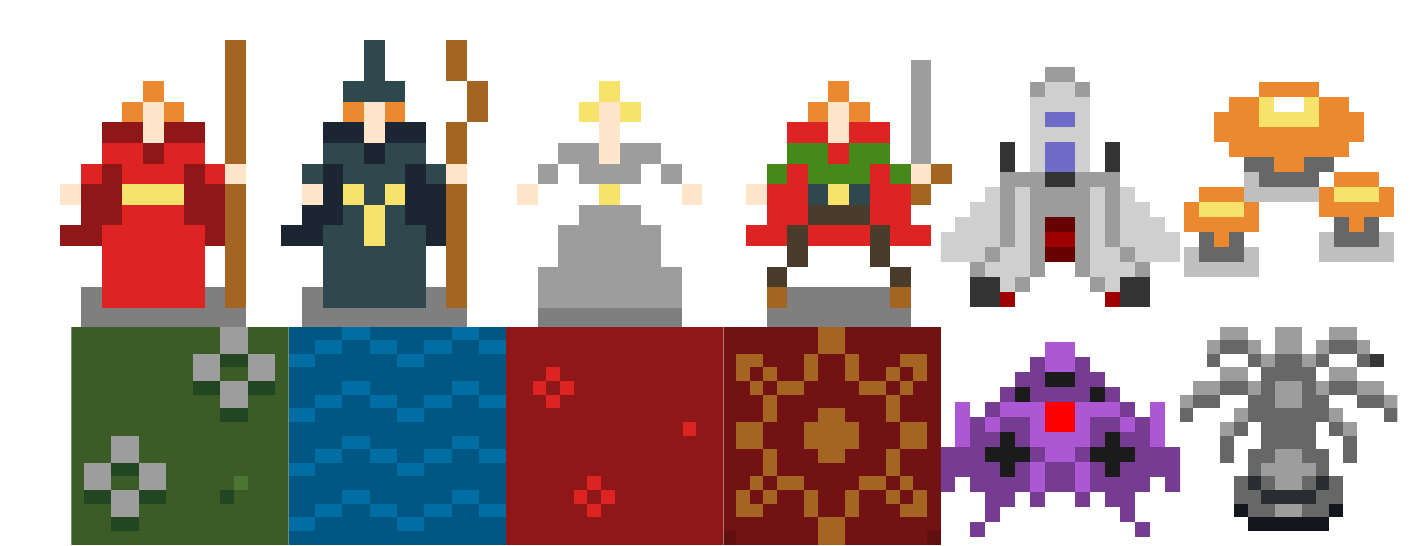}
  \caption{An illustration of tiles from Oryx Design Lab.}
    \label{fig:tiles}
\end{figure}
An important aspect of our world's creation lies in selecting tiles. We use Oryx Design Lab's\footnote{\url{https://www.oryxdesignlab.com/}} tilesets to create two datasets to retrieve tiles. As mentioned above, the first dataset is for environment and interactive/object tiles and the second dataset is for characters. Datasets are represented by an image of the tile or sprite and a description of what the tile is. We label the tile ourselves to increase the effectiveness of the retrieval system. We have $473$ rows for the environment and interactive/object tile dataset and $213$ rows for the character tile dataset. One tile can be repeated for more than one description. In our PCG pipeline, we use DistilBERT~\cite{sanh2019distilbert} to convert the descriptions in the dataset and the description of the tile in the world to the embeddings. Thereafter, we use cosine similarity~\cite{rahutomo2012semantic} to find the most similar tile.  

\subsection{LLM Agent}
\begin{figure}[H]
    \centering
    \includegraphics[width=1.0\linewidth]{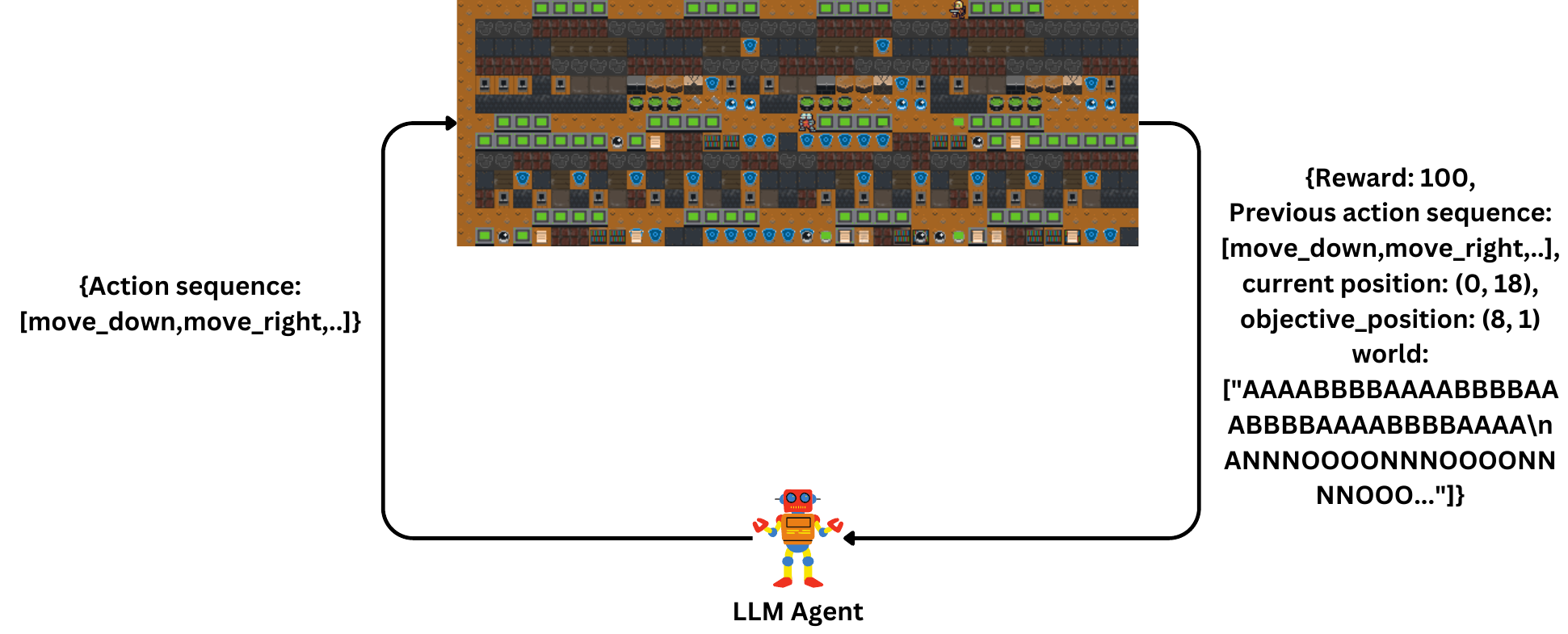}
  \caption{Illustration of how an LLM agent works. LLM agent receives the world, its position, the position of the objective, previous action sequence, and reward for the previous objective. This loop continues till the agent has generated action sequences for all objectives.}
    \label{fig:llm_agent}
\end{figure}

In previous years we have witnessed immense ability of LLMs solving problems in zero-shot and few-shot attempts. We show that LLMs have zero-shot and few-shot ability to plan and traverse in a 2D game as well. We use LLMs as learnt agents trying to traverse towards the objective and complete the objective. As Figure \ref{fig:llm_agent} shows that the LLM agent can be set up in a similar loop to the Reinforcement Learning loop with a world model. The sequence of actions generated by the LLM is the combination of the following actions: \textit{move up, move down, move right, move left, pick object, hit enemy}. LLM receives the location of the player, the location of the object, the world, and is asked to generate the sequence for the first objective. For the next objective and so on, LLM receives the last generated actions and the rewards as well. Maximum reward is for completing the object, then rewards are for reaching as close to the objective as possible, and lastly being far away would cause regret.  



\begin{figure*}[t]
    \centering
    \includegraphics[width=1\linewidth]{Figures/example_7.pdf}
  \caption{An illustration of an example of the complete pipeline. Story is provided in each step of information extraction. All previous information is provided when a piece of information is being extracted, for example, to extract walkable tiles, we pass story, tile mapping, goals, and character information. }
    \label{fig:example}
\end{figure*}

\section{Evaluations}

Evaluations are the key aspect of any AI system. For \texttt{Word2World}, we are interested in evaluating how coherent the world is with the story. We are also interested in finding out if \texttt{Word2World} can create diverse and playable worlds, which can also be used for research purposes. Thus, we use evaluations through two modes: LLM-based evaluations, and conventional PCG checks. The following are the evaluations we ask LLM, as a prompt, by giving the following questions:
\begin{itemize}
    \item \textbf{LLM agent rewards}: Is the world easy enough to be solved?
    \item \textbf{Coherence}: One of the most important metrics. We look for coherence between the story and the world. We provide LLM with the story, the tile mapping and the world, and ask it to give a score up to 100, where 100 is the highest coherence. We ask the LLM to judge based on the coherence between the environment story is setting in and the environment of the generated world, if the characters are similar in the story and the world, and if the objectives for the protagonist are coherent with the story. 
\end{itemize}
Following are the evaluations done by conventional PCG methods:
\begin{itemize}
    \item \textbf{Playability}: We use the AStar agent to find the paths to complete all objectives.
    \item \textbf{Path length}: We use the AStar agent to find the shortest path to achieve all objectives.
    \item \textbf{Novelty}: We measure novelty by the euclidean distance between the worlds generated in each round. We set a threshold for a minimum euclidean distance needed for a world to be called novel. The world should be novel from all worlds created in previous rounds.
    \item \textbf{Novel and Playable}: A world is first checked for \textit{Novelty}, if it passes the check then it is checked for \textit{Playability}.
    \item \textbf{Completion}: \textit{Completion} is defined as true if the whole run of \texttt{Word2World} is completed within 10 tries across the whole pipeline.
    \item \textbf{Accuracy of character tiles}: \[\frac{No.\;of\;characters\;placed}{No.\;of\;characters\;in\;the\;story}\]
    \item \textbf{Accuracy of important tiles}: \[\frac{No.\;of\;important\;tiles\;placed}{No.\;of\;important\;tiles\;in\;the\;story}\] where \textit{important tiles} include environment and interactive tiles that are necessary for the game.
\end{itemize}



\section{Experiments} \label{experiments}

\newcommand{\rrr}[1]{0.4\linewidth}
\begin{figure*}
    \centering
    \begin{subfigure}{\rrr}
        \includegraphics[width=1.01\textwidth]{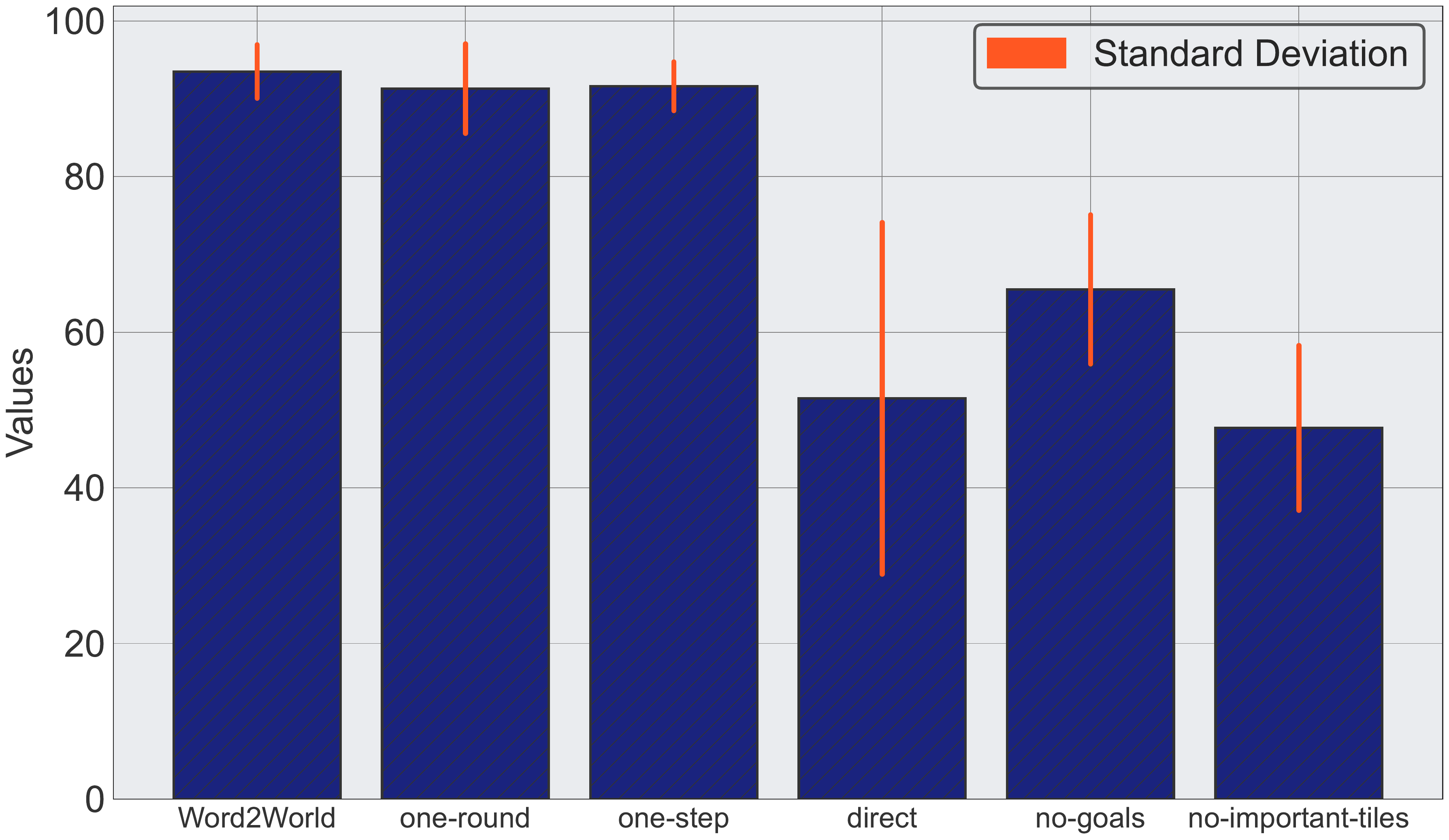}
        \caption{Coherence}
        \label{coh}
    \end{subfigure}
    \begin{subfigure}{\rrr}
        \includegraphics[width=1.01\textwidth]{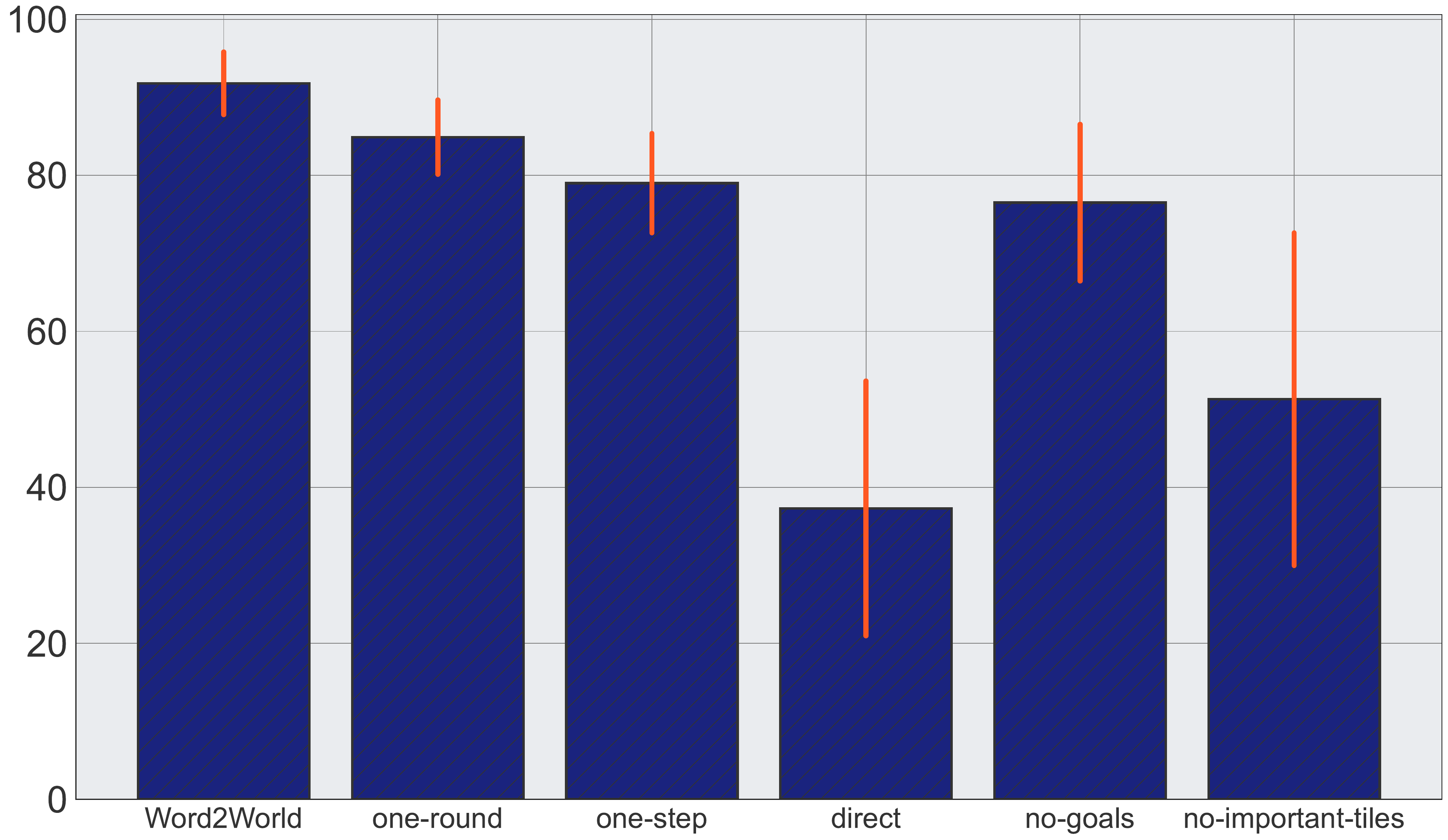}
        \caption{Acc. of character tiles}
        \label{achar}
    \end{subfigure}
    \begin{subfigure}{\rrr}
        \includegraphics[width=1.01\textwidth]{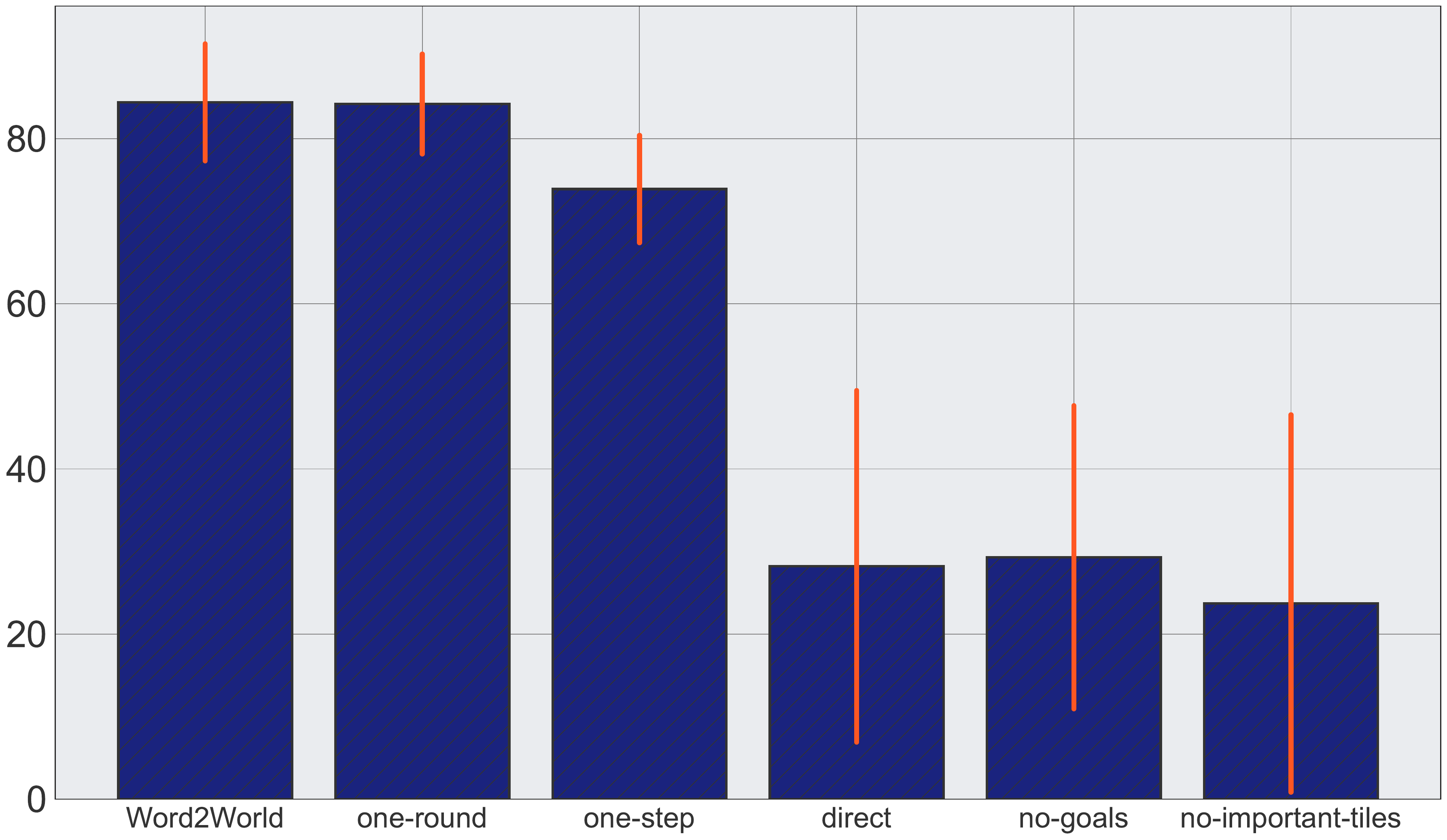}
        \caption{Acc. of important tiles}
        \label{aimp}
    \end{subfigure}
    \begin{subfigure}{\rrr}
        \includegraphics[width=1.01\textwidth]{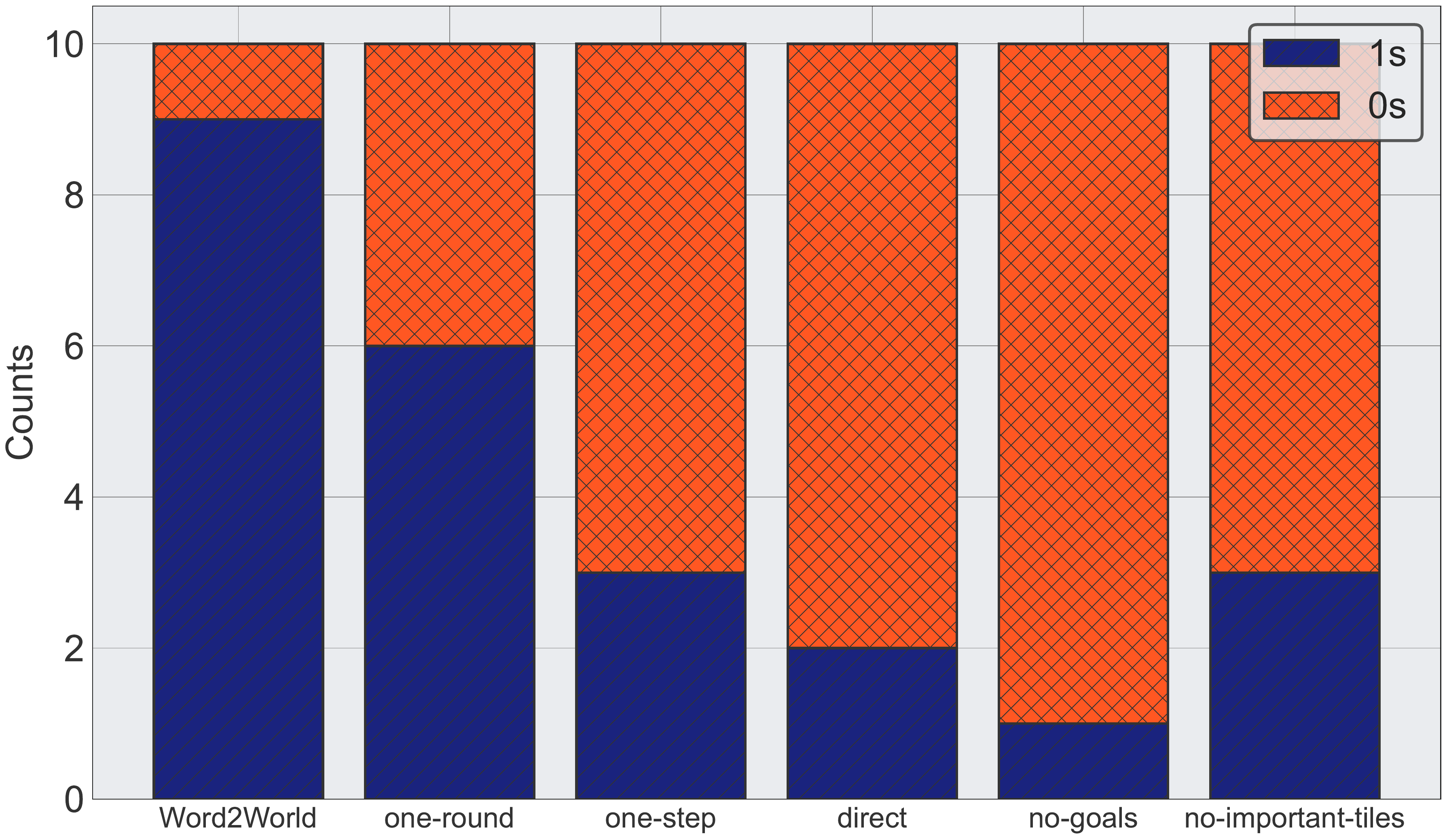}
        \caption{Playability}
        \label{pla}
    \end{subfigure}
    \caption{A comparison of ablation studies with respect to (a) Coherence, (b) Playability, (c) Accuracy of character tiles, and (d) Accuracy of important tiles. From left to right in each image the bars represent \textit{Word2World, one-round-generation, one-step-generation, direct-generation, no-goals-extracted} and \textit{No-important-tiles-extracted}.}
    \label{fig:ablate}
\end{figure*}

To conduct our experiments, we complete 10 runs to show the statistical significance of our experiments. For our hyperparameters for the story, we set 4-5 story paragraphs and 8 objectives for the protagonist to complete the game. For the creation of the world, we use a maximum of 15 important tiles to be placed. We set this parameter by experimenting across different ranges of parameters and found these to be the best set of parameters. We will discuss the reason behind it in the Results section.

 We compare \textit{claude-3-opus-20240229}, \textit{claude-3-sonnet-20240229}, \textit{claude-3-haiku-20240307}, \textit{gpt-4-turbo-2024-04-09}, \textit{gpt-4-0613}, and \textit{gpt-3.5-turbo-0125} on \textit{Playability}, \textit{Novelty}, \textit{Novel and Playable}, \textit{Coherence}, \textit{LLM Agent Rewards}, and  \textit{Completion}. The experiments are again conducted for 10 runs. All metrics are counts while LLM Agent Rewards is the actual normalised reward gathered after 2 episodes. An episode is considered finished when LLM has generated action sequences for all objectives.
 
 We set the temperature to be 1 across all steps. We give a maximum of $1000$ iterations for the Astar agent to find paths to all objectives.

As our approach is a novel text-to-environment system, we do a thorough ablation study to observe if all steps are necessary. Thus creating the following versions of our system to do the ablation study:

\begin{itemize}
    \item \textit{direct-generation}: Generating directly from the story. No steps were implemented in between.
    \item \textit{no-goals-extracted}: Generation without extraction of goals.
    \item \textit{No-important-tiles-extracted}: Generation without extraction of walkable, objective and important tiles.
    \item \textit{one-step-generation}: Generating environment, character and interactive tiles in one go, and not in 2-steps, like \texttt{Word2World}.
    \item \textit{one-round-generation}: Generation with single round.
\end{itemize}

\section{Results}

In this section, we will explore the results of our experiments, that were defined in the previous section. We will dive deep into each step, looking into how our system produces the world. We will then look into the results of ablations studies.

\subsection{Pipeline Exploration}

To show how each step in Figure \ref{fig:flowchart} works, we go through an example of a world generation in Figure \ref{fig:example}. The generated story in Figure \ref{fig:example} is as follows:\\ \textit{In a world where the ever-stretching, dense forests of Eldoria whispered ancient secrets and the lively rivers danced with untamed vigor, there lived a young archer named Lyra. Her home, a picturesque village nestled within the heart of these forests, thrived on harmony and a deep connection to nature. However, peace was threatened when an ominous shadow fell over the village, cast by a sorceress named Vespera, who sought to harness the ancient magic hidden beneath the village to bend the natural world to her will.\\ Lyra, with the spirit of the forest running vigorous in her veins, took it upon herself to save her home and the balance of nature. Her objectives were clear and daunting; locate the ancient artifacts scattered through the lands that could negate the sorceress's dark magic, rally the scattered tribes of Eldoria for strength in unity, master the elemental archery taught by the mythical guardians, seek a legendary elixir to shield her from Vespera's curses, decipher the ancient texts revealing Vespera's weaknesses, sabotage the sorceress's outposts to slow her advance, rescue the creatures ensnared by dark magic to gain allies, and ultimately, face Vespera in a decisive battle to protect the sanctity of her home.\\ As her journey unfolded, Lyra traversed through the enchanted groves where the trees sang melodies of ancient times, scaled the daunting Silverpeak mountains where the winds whispered secrets of might, and delved into the shadowy depths of the Whispering Caves, where the line between reality and illusion blurred. Each step was a battle against the forces marshaled by Vespera, who, in her own twisted way, believed she was bringing a new order that would elevate her to goddess status.\\ Vespera, aware of Lyra's quest, unleashed her minions and used her dark magic to turn the very environment against Lyra. She summoned storms that roared with anger, twisted the creatures of the forest into monstrous versions of themselves, and laid traps that blurred the lines between the physical and mystical worlds. Yet, with every challenge, Lyra's resolve hardened, her skills sharpened, and alliances strengthened.\\ The climax arrived as Lyra, armed with the artifacts and bolstered by the forces of Eldoria, stood at the precipice of Vespera's sanctum. The battle was fierce, a tempest of magic, arrows, and the clashing wills of two indomitable spirits. In the end, it was Lyra's purity of heart and unwavering determination that pierced through the darkness. With Vespera's defeat, the corrupt magic dissipated, and nature reclaimed what was lost. The victory was not just a testament to Lyra's bravery but to the enduring spirit of Eldoria. Peace returned to the village, and the forests sang louder than ever, a hymn of harmony between the natural world and its guardians.} 

It can be observed that the story defines a protagonist, an antagonist and some objectives. From the story, we extract brief descriptions of characters as we want to find similarities between these descriptions and the descriptions of characters available in our dataset. Next, we extract tile mapping. This requires us to provide story and character information. From tile mapping, we can observe that it is coherent with the story. We have all the characters and objects extracted. We have tiles related to village, such as grass, dirt path, bushes, stream, wooden bridge etc. as described in the story. 

The story is set in a forest, so it would have been better to have a tree tile as well. Also, tiles like treehouse and wooden bridge will have scaling problems. We will discuss this in more detail in the Discussion section when we discuss the limitations of \texttt{Word2World}. Once we extract goals and tiles needed to create a game, we prompt LLM to extract important, walkable, and interative tiles. in Figure \ref{fig:example}, we can observe that these tiles can help with defining constraints. After all the information extraction, LLM is prompted to create the world without the characters and then again prompted to place the characters and interactive objects. Hereafter, we ask LLM to find the position of all the objectives. From the example, we can see that position of objectives are correctly extracted by the LLM. The position of objectives also shows us that the coherence between the created world, goals, tile mapping, and the story is correct, as all of the extracted information resonates with the story. Through a similarity score calculated by distilBERT, we finally get the world in a top-down 2D setting. 

\begin{table*}[]
\centering
\caption{A comparison between family of gpt and claude models. All metrics are explained in Section \ref{experiments}.}
\label{tab:comp}
\begin{tabular}{|l|l|l|l|l|l|l|}
\hline
Model                           & Novelty & Playability & Novel and Playable & Coherence & LLM Agent Rewards & Completion \\ \hline
\textit{claude-3-opus-20240229}          & $\boldsymbol{9}$    & $8$        & $6$                  & $\boldsymbol{10}$       & $\boldsymbol{0.3421 \pm 0.37}$ &$ $7          \\ \hline
\textit{claude-3-sonnet-20240229}        & $7$     & $7$        & $5$                  & $7$         & $0.2444 \pm 0.21$ & $7$          \\ \hline
\textit{claude-3-haiku-20240307 }        & $7$     & $5$        & $4$                  & $7$         & $0.2093 \pm 0.21$ & $5$          \\ \hline
\textit{gpt-4-turbo-2024-04-09} & $8$     & $\boldsymbol{9}$        & $\boldsymbol{7}$                  & $9$         & $0.3285 \pm 0.16$ & $\boldsymbol{10} $        \\ \hline
\textit{gpt-4-0613}             & $6$     & $8$        & $5$                  & $8$         & $0.2583 \pm 0.08$ & $7$          \\ \hline
\textit{gpt-3.5-turbo-0125}     & $6$     & $7$        & $5$                  & $5$         & $0.1905 \pm 0.19$ & $5$          \\ \hline
\end{tabular}
\end{table*}

\subsection{Ablation Study Results}
\begin{figure}
    \centering
    \includegraphics[width=1.0\linewidth]{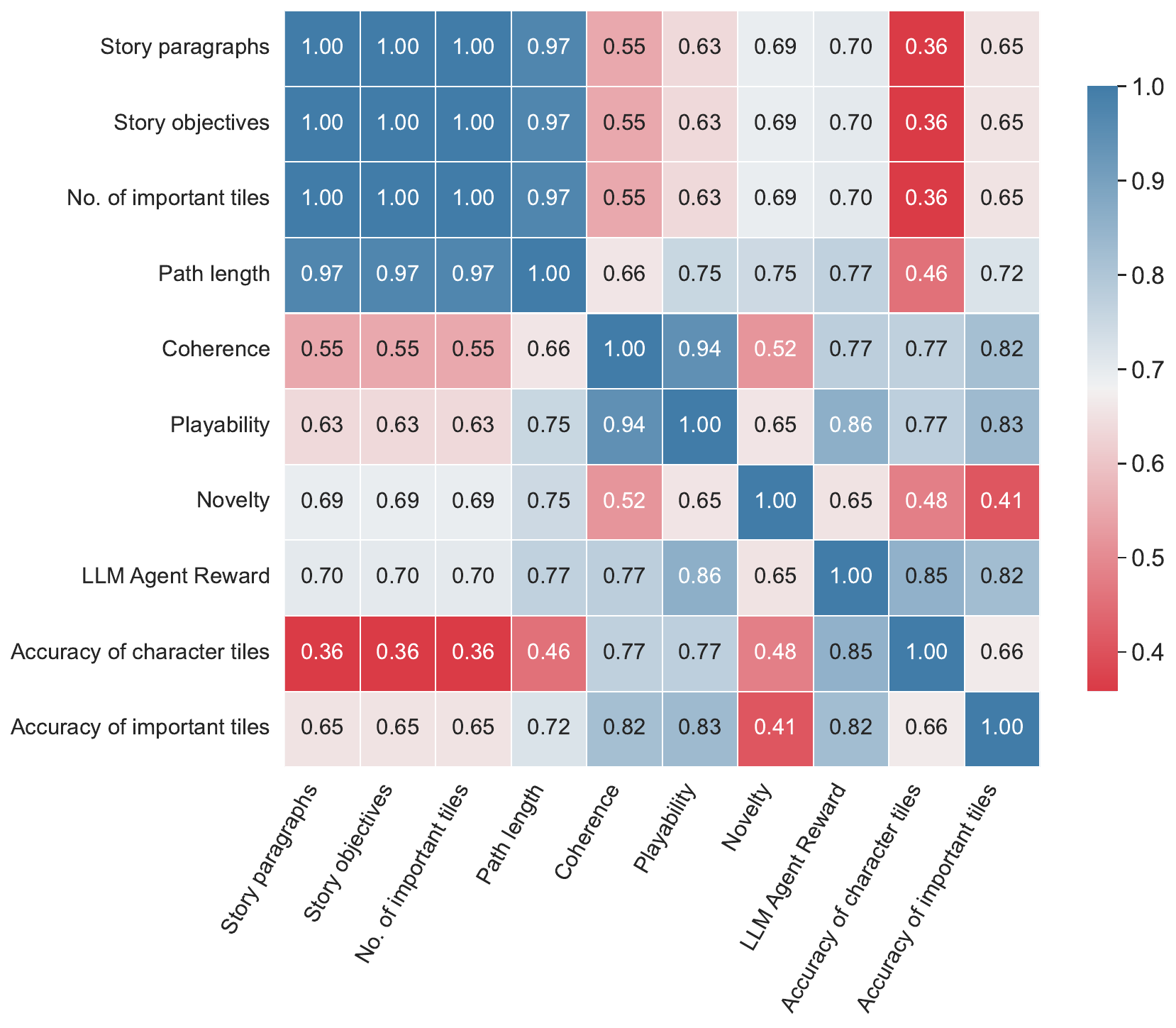}
  \caption{Correlation between world generation parameters, that are \textit{Story Paragraphs}, \textit{Story objectives} and \textit{No. of important tiles} and evaluation metrics. Note that Path length is highly correlated and metrics generated by GPT-4 are usually less correlated to the world generation parameters.}
    \label{fig:corr}
\end{figure}
We will start with one of the most important results: \textbf{Coherence} in Figure \ref{coh}. This metric tells us about how coherent the world is with the story. We can observe that our method, \texttt{Word2World}, generated the most coherent worlds. This is intuitive as each step in our method may act as a stepping stone, and providing the extracted information back to the LLM may give it more planning ability. Furthermore, one-round-generation and one-step-generation also produce coherent results as all other steps are included, giving the LLM enough information to reason. On the other hand, direct-generation acts as a random generator, as there is no context other than the story itself, which may provide only some context for the reasoning ability of the LLM.  

Figure \ref{achar} shows a comparison of ablation studies on \textbf{Accuracy of character tiles}. Here, \texttt{Word2World}, one-round-generation and one-step-generation produce similar results while goal extractions seem to have less impact on the pipeline when placing characters. Extraction of important tiles makes a substantial difference in the results, having a mean score of 50. The difference between \texttt{Word2World} and irect-generation tells us that the whole pipeline is necessary to place all the characters. \textbf{Accuracy of important tiles}, in Figure \ref{aimp}, shows us that \texttt{Word2World} with three rounds of generation and one round of generation have an almost negligible difference, and one-step-generation is also close to them, but all other ablations show poor results, through which we can conclude that all steps are required to achieve higher accuracy of important tiles being placed. 

\textbf{Playability}, in Figure \ref{pla}, shows that 9 out of 10 times \texttt{Word2World} can produce playable worlds. one-round-generation will get worse and generate 60$\%$ playable levels. This result shows that rounds are important to generate playable worlds. Poor results of one-step-generation tell us that placing environment and character tiles separately is a very important step. The rest of the ablations give us poor results, and intuitively, if goals are not extracted then playability becomes very low. Thus, this result confirms that extracting goals is an important step.

\subsection{Other Results}

\textbf{Correlation Between Input And Output:} In Figure \ref{fig:corr}, we observe the correlation of world generation parameters to the evaluation metrics. \textit{Story Paragraphs}, \textit{Story objectives} and \textit{No. of important tiles} are the World Generation Parameters(WGPs). This result is produced by varying the number WGPs. The figure shows an overall high correlation with each other. \textit{Path leangth} is highly correlated to WGPs, which means that we can get larger path lengths if we increase the values of WGPs. With this result, we can also say that we can achieve complex environments procedurally by increasing the values of WGPs incrementally. Most of the evaluation metrics are highly correlated to WGPs, except \textit{Coherence} and \textit{Accuracy of character tiles}. This means that a change of value in WGP will not necessarily mean that we will get high coherence between the story and the generated world. Similarly, the accuracy of character tiles is also not ensured by the varying WGPs.

\begin{figure}
    \centering
    \includegraphics[width=1\linewidth]{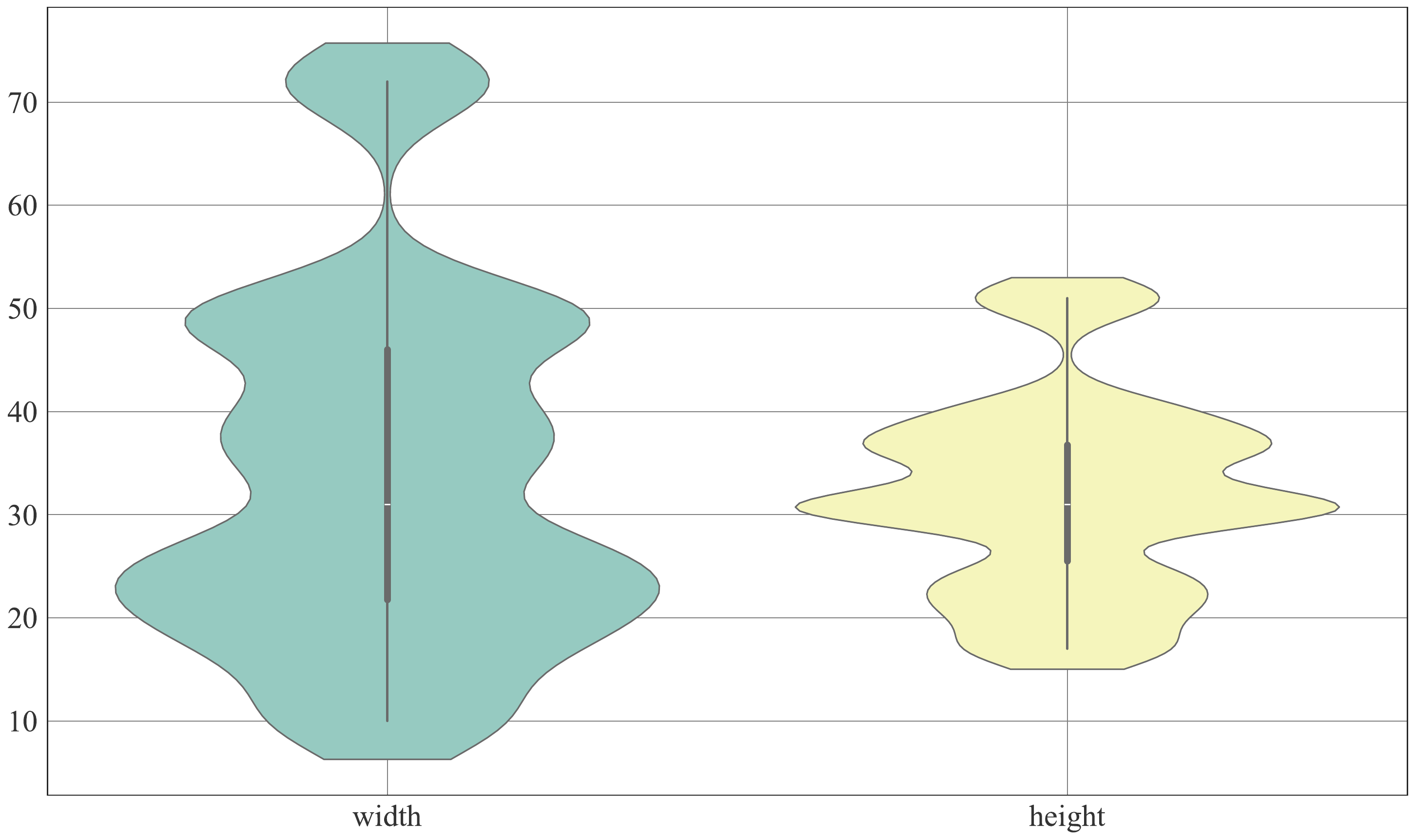}
  \caption{Illustration of how wide the world map size distribution is. This illustration shows that our technique can create different size world maps.}
    \label{fig:violin}
\end{figure}

\textbf{Diverse Worlds:} Another important result is illustrated in Figure \ref{fig:violin}. This result shows that \texttt{Word2World} can produce worlds with varying sizes. This also shows us that \texttt{Word2World} has a larger distribution when it comes to the width of the world, compared to the height. This also proves that \texttt{Word2World} can endlessly create worlds. Furthermore, the diversity of worlds is illustrated in Figure \ref{fig:world_examples}. We can observe that the environment tiles are placed with many diverse patterns. 

\begin{figure}[h!]
    \centering
    \includegraphics[width=1.0\linewidth]{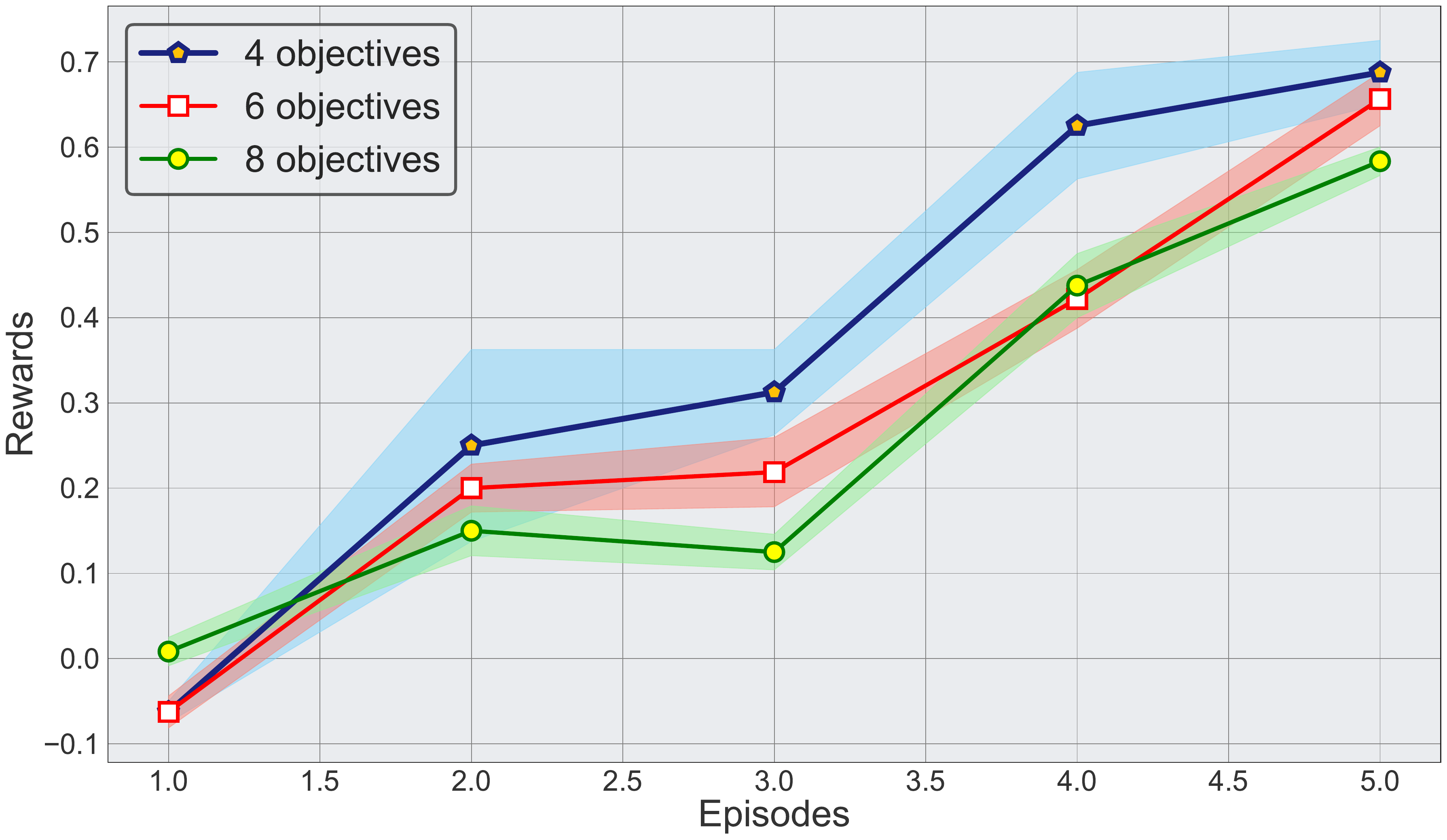}
  \caption{LLM Agent Rewards per episode. One episode is considered finished when LLM has generated sequences of actions for all the objectives. We compare on worlds with the 4, 6 and 8 objectives. The rewards are normalised. We compare for 5 episodes. Each experiment spans 10 seeds. The shaded region is the standard deviation and the line represents the mean value.}
    \label{fig:rewards_per_episode}
\end{figure}

\textbf{LLM Agent Rewards:} Figure \ref{fig:rewards_per_episode} shows an important result where we compare the LLM Agent Rewards on 3 different worlds: a world with 4 objectives, 6 objectives, and 8 objectives. We show that the LLM's capability of generating action sequences that result in high reward increases the episodes increase. This is due to the fact that LLM receives feedback with the previous generated sequence for the all previous episodes and the rewards for it. This shows that with enough context, LLM can act as a learnt agent and produce high-rewarding sequences.

\textbf{LLM Comparisons:} Table \ref{tab:comp} compares the family of GPT and Claude models. We take into account \textit{Novelty, Playability, Novel and Playable, Coherence, LLM Agent Rewards} and \textit{Completion} into account. We observe that the largest models in both families perform the best. \textit{claude-3-opus-20240229} performs the best on \textit{Novel}, \textit{Coherence}, and \textit{LLM Agent Rewards}, while \textit{gpt-4-turbo-2024-04-09} performs the better than all other compared models on \textit{Playability}, \textit{Novel and Playable} and \textit{Completion}. From the results in Table \ref{tab:comp}, we can deduce that larger language models may overall perform better. Reasoning and planning abilities can be judged by metrics like \textit{Coherrence} and \textit{LLM Agent Rewards}, where \textit{claude-3-opus-20240229} excels. LLMs capability to extract the right information and adhere to the prompt can be judged by \textit{Completion} where \textit{gpt-4-turbo-2024-04-09} performs better. \textit{Novelty}, \textit{Playability}, and \textit{Novel and Playable} can be used to judge how expressive but controllable an LLM is. Here, \textit{gpt-4-turbo-2024-04-09} is slightly better than \textit{claude-3-opus-20240229} by creating more novel and playable worlds. Overall, \textit{claude-3-opus-20240229} and \textit{gpt-4-turbo-2024-04-09} perform almost similar across all metrics and much better than their smaller versions.

\section{Related work}

\texttt{Word2World} is a system that starts from a prompt to generate a story and then extracts all the relevant information needed to create a playable game. The generation of games from stories has been researched in previous works. \citet{ammanabrolu2020bringing} used knowledge graphs to place objects in the world generated from a story. \citet{fan2020generating} uses Light~\cite{urbanek2019learning}, a dataset that contains information about objects, characters and locations of a fantasy world, to train a neural network to place objects in an interactive world. Furthermore, \citet{hartsook2011toward} used search-based optimisation~\cite{togelius2011search} to generate games from stories through a graph which dictates where an object is placed. \texttt{Word2World} differs from all of the previous work as it generates all the components illustrated in Figure \ref{fig:flowchart}, thus having diversity in the generation of all components.

Our work falls under Procedural Content Generation (PCG)~\cite{shaker2016procedural} that has found great success through machine learning algorithms, by the use of deep learning and evolutionary algorithms~\cite{togelius2011search, summerville2018procedural, liu2021deep, beukman2023hierarchically}. Prior work in PCG has demonstrated that level generation can be designed as a sequence generation problem, thus using sequence-based prediction models, such as LSTMs, can effectively generate levels~\citep{summerville2016super}. After the advancements in Transformer-based LLMs~\citep{vaswani2017attention}, LLMs have shown to effectively generate diverse sets of levels through supervised learning where ~\citet{todd2023level} has generated Sokoban~\cite{murase1996automatic} levels through GPT-2 and GPT-3, while ~\citet{sudhakaran2024mariogpt} generated Super Mario bros. levels~\citet{togelius2011procedural} through GPT-2. Furthermore, ~\citet{nasir2023practical} has used the human-in-the-loop and bootstrapping~\cite{torrado2020bootstrapping} method to generate diverse and playable levels from scarce human-designed data using GPT-3. \citet{bruce2024genie} introduced Genie, which was trained on unlabeled internet videos, to produce games via text prompts and images. Recent work also shows that LLMs can be integrated to improve or create game mechanics~\cite{anjum2024ink, hu2024generating}. We introduce a pipeline that uses LLM to create complex levels (or worlds) in a few-shot manner.

\texttt{Word2World} can be considered as a Game Master (GM) that creates a story and then places all the components of the story in a top-down role-playing game, for the player to play. A Game Master (GM) is a person who creates the plot, characters, and narrative of the game~\cite{gallotta2024large}. The first notable GM was \textit{AI Dungeon}~\cite{hua2020playing}, which was managed by a fine-tuned version of GPT-2. The goal of AI Dungeon is to interact with humans and then progress the story of the game with respect to the input received from humans. \citet{zhu2023calypso} introduced \textit{CALYPSO} that helps a GM brainstorm ideas and presents texts that can be displayed directly to the players. \citet{callison2022dungeons} trained an LLM on 900 games to act as a GM.

As \texttt{Word2World} is a system that designs the whole game from scratch, it can be placed in a multi-faceted game generation. \citep{cook2011multi} introduced ANGELINA which used an evolutionary algorithm to create puzzle games. Later on, ANGELINA was extended to platformer games as well \cite{cook2012angelina, cook2012aesthetic}. Furthermore, we put \texttt{Word2World} under Computational Game Creativity \cite{browne2012computational,liapis2014computational} as game generation is a multi-faceted generation, each facet requires creativity. For example in \texttt{Word2World}, the sotry, character descriptions, tile descriptions, goal descriptions, and world generation, each of these steps requires computational creativity   

\texttt{Word2World} uses an LLM agent to traverse through the world, trying to achieve objectives. The LLM agent is used as an evaluator to see if an agent can traverse the world. Previous works show that some LLMs can play chess~\cite{noever2020chess} and Go~\cite{ciolino2020go}. ~\citep{reed2022generalist} introduced GATO, a generalist agent that can perform many tasks, including playing various Atari games and achieving near-human level scores. ~\citep{de2024will} has also shown that GPT-4 can play Doom, though it is a slow pipeline as it needs many iterations of prompting to complete the game. It is to be noted that LLM-based agents require finely tuned prompts to play games and all LLMs are not capable of playing games.

\newcommand{\kkk}[1]{0.3\linewidth}
\begin{figure*}
    \centering
    \begin{subfigure}{\kkk}
        \includegraphics[width=1.01\textwidth]{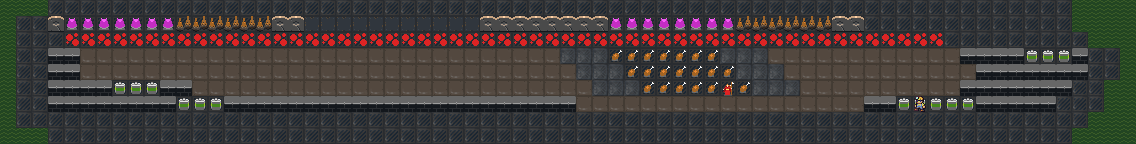}
        \caption{}
        \label{d}
    \end{subfigure}
    \begin{subfigure}{\kkk}
        \includegraphics[width=1.01\textwidth]{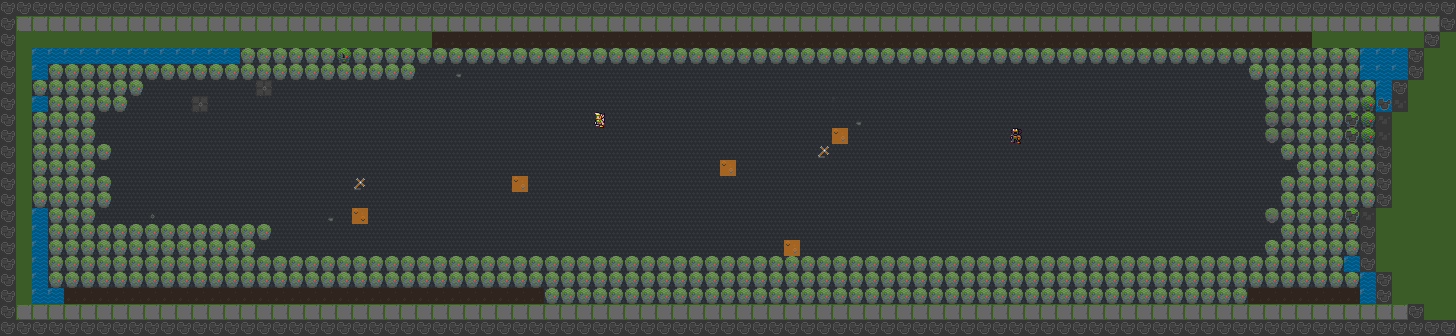}
        \caption{}
        \label{b}
    \end{subfigure}
    \begin{subfigure}{\kkk}
        \includegraphics[width=1.01\textwidth]{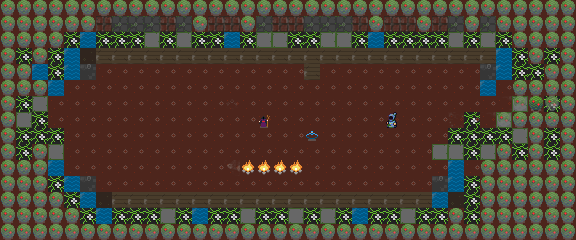}
        \caption{}
        \label{a}
    \end{subfigure}
    \begin{subfigure}{\kkk}
        \includegraphics[width=1.01\textwidth]{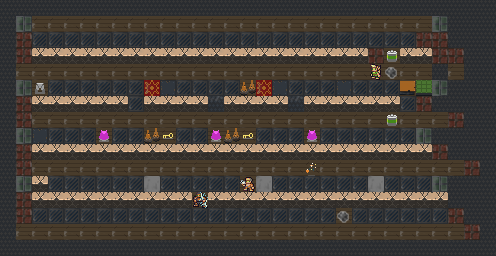}
        \caption{}
        \label{c}
    \end{subfigure}
    \begin{subfigure}{\kkk}
        \includegraphics[width=1.01\textwidth]{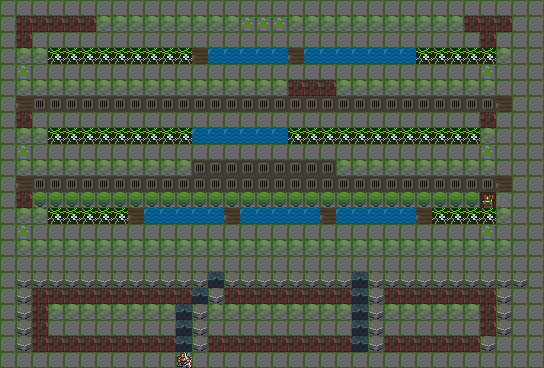}
        \caption{}
        \label{f}
    \end{subfigure}
    \begin{subfigure}{\kkk}
        \includegraphics[width=1.01\textwidth]{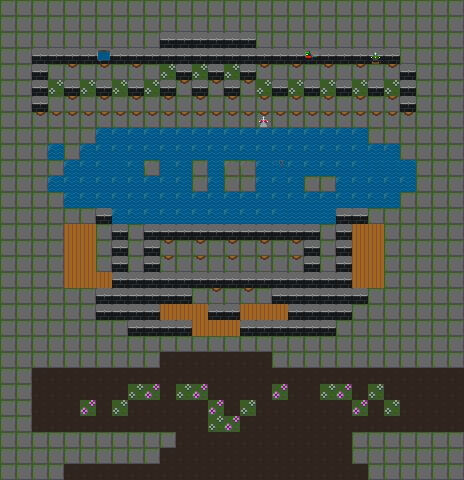}
        \caption{}
        \label{e}
    \end{subfigure}
    \caption{Examples of the diversity in the generated worlds.}
    \label{fig:world_examples}
\end{figure*}

\section{Discussion}

In this section, we will discuss the limitations of \texttt{Word2World}. Following are the limitations we observed:

\begin{enumerate}
    \item \texttt{Word2World} is designed only for top-down games.
    \item Expressiveness of \texttt{Word2World} is limited to the tile dataset it retrieves the tiles from.
    \item \texttt{Word2World} does not handle tile scaling issues. For example, a house is not semantically placed over more than one tile. This leads to limitations in the expressiveness and aesthetics of the world.
    \item There is no mechanism for animation in \texttt{Word2World}.
    \item Game mechanics are predefined and set for all games.
    \item Expressiveness of \texttt{Word2World} is limited by the capability of underlying LLM.
    \item As an open-sourced LLM is not used, one run of \texttt{Word2World} can cost around $\$0.5$ to $\$1$ per generation based on the LLM used and the tokens generated. 
    \item As an open-sourced LLM is not used, \texttt{Word2World} can not run fully local hardware.
    \item Stories generated follow almost the same set of events placed in different environments.
    
\end{enumerate}

\section{Conclusion}

To conclude, we introduced \texttt{Word2World}, a novel LLM-based pipeline to generate playable worlds from stories. \texttt{Word2World} leverages the LLM capabilities of generating diverse content and extracting information from content. Though LLMs tend to uncontrollably generate content, \texttt{Word2World} takes that into account and generates controllably. This is proven by \texttt{Word2World} generating playable levels $90\%$ of the time.  We also show that \texttt{Word2World} can create worlds that are highly coherent with the story. 

We believe this direction of the research is highly important for the research community as well as the industry. \texttt{Word2World} can act as a narrative generation to drafting levels for the industry, while for the research community it can provide diverse environments for many reinforcement learning paradigms, such as Open-ended learning where the system requires an environment generator to produce highly-diverse environments. We have empirically proved that \texttt{Word2World} can generate highly diverse environments. We believe that there are many future directions for \texttt{Word2World}. It can be extended to many different settings of Role-playing Games (RPG), such as 2D platformer or 3D. It can be extended to create open-world games for storybooks. As \texttt{Word2World} has the ability to create highly diverse worlds, these worlds can act as environments for Reinforcement Learning~\cite{sutton2018reinforcement} agents and can help in an open-ended learning~\cite{wang2019paired, jiang2021replay, nasir2023augmentative} setting.

\bibliography{main}

\end{document}